\documentclass[a4paper,UKenglish,cleveref, autoref, thm-restate]{lipics-v2021}

\pdfoutput=1 
\hideLIPIcs  

\usepackage{amsfonts}       
\usepackage{nicefrac}       
\usepackage{microtype}      

\usepackage[table,xcdraw]{xcolor}
\usepackage{amsfonts}
\usepackage[misc]{ifsym}
\usepackage{amsmath}
\usepackage{booktabs}
\usepackage{amsthm}
\usepackage{mathtools}
\usepackage{thmtools}
\usepackage[ruled,linesnumbered,noend]{algorithm2e}
\usepackage{lipsum,graphicx,subcaption}

\newtheorem*{pfs*}{Proof Sketch}
\newtheorem*{proof*}{Proof}
\usepackage[noend]{algorithmic}

\DeclareMathOperator*{\argmax}{arg\,max}

\usepackage{tcolorbox}
\usepackage{physics}
\usepackage{todonotes}

\title{DPMS: An ADD-Based Symbolic Approach for Generalized MaxSAT Solving}

\author{Anastasios Kyrillidis}{Department of Computer Science, Rice University, USA}{}{}{}

\author{Moshe {Y. Vardi}}{Department of Computer Science, Rice University, USA}{}{}{}

\author{Zhiwei Zhang \footnote{The author list has been sorted alphabetically by last name; this should not be used to determine the extent of authors’ contributions. Corresponding author: Zhiwei Zhang (zhiwei@rice.edu).}} {Department of Computer Science, Rice University, USA}{zhiwei@rice.edu}{}{}

\authorrunning{A. Kyrillidis, M. Vardi and Z. Zhang} 



\category{} 

\relatedversion{} 

\nolinenumbers 

\EventNoEds{2}
\EventShortTitle{Arxiv}
\EventAcronym{Arxiv}
\SeriesVolume{42}
\ArticleNo{}

\begin{document}

\maketitle

\begin{abstract}
Boolean MaxSAT, as well as generalized formulations such as Min-MaxSAT and Max-hybrid-SAT, are fundamental optimization problems in Boolean reasoning. 
Existing methods for MaxSAT
have been successful in solving benchmarks in CNF format. 
They lack, however, the ability to  handle 1) (non-CNF) hybrid constraints, such as XORs and 2) generalized MaxSAT problems natively. 
To address this issue, we propose a novel dynamic-programming  approach for solving generalized MaxSAT problems with hybrid constraints --called \emph{Dynamic-Programming-MaxSAT} or DPMS for short-- based on Algebraic Decision Diagrams (ADDs). 
With the power of ADDs and the (graded) project-join-tree builder, our versatile framework admits many generalizations of CNF-MaxSAT, such as MaxSAT, Min-MaxSAT, and MinSAT with hybrid constraints.
Moreover, DPMS scales provably well on instances with low width. 
Empirical results indicate that DPMS is able to solve certain problems quickly, where other algorithms based on various techniques all fail. Hence, DPMS is a promising framework and opens a new line of research that invites more investigation in the future. 
\end{abstract}
\section{Introduction}
The \emph{Maximum Satisfiability} Problem (MaxSAT) is the optimization version of the fundamental Boolean satisfiability problem (SAT). 
MaxSAT asks for the maximum number of constraints (maximum total weight in  weighted setting) that can be satisfied simultaneously by an assignment. 
For each assignment, the number (total weight, in the weighted setting) of violated constraints is often regarded as the \emph{cost} that should be minimized. 
MaxSAT finds abundant applications in many areas \cite{applicationOfMaxSAT}, including  scheduling \cite{scheduling}, planning \cite{planning}, 
software debugging \cite{debugging}, hardware security \cite{zhang2020automated} and explainable AI (XAI) \cite{sakai2020bnn} .
Ideally speaking, the constraints may be of any type, though formulas in conjunctive normal form (CNF) are most studied both in theory and practice \cite{CNFEncodings}. 
In spite of the NP-hardness of MaxSAT, there has been dramatic progress on the engineering side of (weighted-partial) Max-CNF-SAT solvers for industrial instances \cite{CDCLSolvers}. 
 
Similar to SAT solvers, MaxSAT solvers can be classified into complete and incomplete ones.
 A complete solver will return an assignment of optimal cost with a guarantee (proof),  while an incomplete algorithm only  gives a ``good" assignment without a guarantee of optimality \cite{incompleteAlgorithms}. Complete (partial) MaxSAT solvers constitute the majority of the MaxSAT solver family. 
Mainstream techniques of complete MaxSAT solvers have been shifting from  DPLL \cite{DPLLMaxSAT} and Branch-and-Bound (B\&B) \cite{branchAndBound} to iterative or core-guided CDCL-based approaches \cite{li2021maxsat,rc2,unsatcoreMaxSAT}. 
Those methods have achieved remarkable performance on large-scale benchmarks in CNF \cite{maxsatCompetition}. Most incomplete MaxSAT solvers are based on local search (LS) and its variants. 
Local search is advantageous for quickly exploring the assignment space and heuristically adapting constraint weights. 
LS-based solvers can be surprisingly efficient in reaching a low-cost region for large instances, which can be infeasible for complete solvers \cite{satlike}. 
While there was a performance gap between complete solvers and LS-based solvers on industrial benchmarks, this gap has been encouragingly squeezed by  recent work \cite{satlike}. Nevertheless, due to the incompleteness of those solvers, they are still not the most preferable solvers in many applications despite their efficiency. 

Regardless of the success of modern Max-CNF-SAT solvers, we point out that their capability of handling more general problems is insufficient. 
In this paper, we consider two important extensions of Max-CNF-SAT. 
The first one is to generalize the type of constraints from CNF clauses to general (\emph{hybrid}) Boolean constraints \cite{fouriersatAIJ} --e.g., cardinality constraints and XORs-- yielding the Max-hybrid-SAT problem. 
Max-hybrid-SAT offers strong expressiveness and numerous applications; e.g., Max-XOR-SAT encodes problems in cryptanalysis \cite{Linear-Cryptanalysis} and maximum-likelihood decoding \cite{maximumLikehoodEncoding}.  
In another direction, there are problems that require Boolean optimization with uncertainty or adversarial agent  cite{QMaxSAT}.
Thus, the second extension is to allow both $\min$ and $\max$ operators to define the Min-MaxSAT problem, whose complexity falls into $\Pi^p_2$-complete, which is likely harder than NP-hard \cite{polynomialHierchary}.
Min-MaxSAT acts as a useful encoding in  combinatorial optimization \cite{minmax} and conditional scheduling \cite{conditionalScheduling}. 

Max-hybrid-SAT and Min-MaxSAT are less well studied compared with MaxSAT.  Furthermore, the dominating techniques in complete modern Max-CNF-SAT solvers (SAT-based, B\&B) are not easily applicable to generalized MaxSAT, due to their dependency on the clausal formulation and the single type of optimization operator ($\max$). 
 Therefore, a versatile framework that handles different variants of MaxSAT is desirable.

In this paper, we propose a novel, versatile, and general framework, called \emph{Dynamic-Programming-MaxSAT} (DPMS). To our best knowledge, DPMS is the first framework that is capable of natively handling all variants of MaxSAT described above. This work is inspired by the success of symbolic dynamic programming in model counting \cite{dpmc}.  
One of the theoretical contributions of our work is to view MaxSAT as \textit{max-of-sum} and borrow ideas from model counting (\textit{sum-of-product}). 
Though \textit{max-of-sum} and \textit{sum-of-product} look different, they are both special cases of  \textit{functional aggregate queries} (FAQs) \cite{faq} and  share common properties such as early projection, which enables us to generalize the approach of \cite{dpmc} to DPMS. 

DPMS consists of two phases: 1) a planning phase where a project-join tree \cite{dpmc} is constructed as a plan and 2) an execution phase, where constraint combinations and variable eliminations are conducted according to this plan. DPMS natively handles a variety of generalized  problems, including weighted-partial MaxSAT, Min-MaxSAT, and Max-hybrid-SAT. Hybrid constraints are handled by DPMS via using \emph{Algebraic Decision Diagrams} (ADDs) to represent  pseudo-Boolean functions, while Min-MaxSAT formulations are handled by using \emph{graded} project-join trees \cite{procount} as plans for execution. Other under-researched formulations such as MinSAT \cite{minsat}  and Max-MinSAT problems can also fit well in our framework. In addition, DPMS explicitly leverages structural information of instances by taking advantage of  the development of tree decomposition tools \cite{flowCutter,tamaki2019positive}. As a result, DPMS scales polynomially on instances with bounded width. 

We developed a software tool implementing the DPMS framework.
We demonstrate in the experiment section that on problems with hybrid constraints and low width, DPMS outperforms start-of-the-art MaxSAT and pseudo-Boolean solvers equipped with CNF encodings. We also show that DPMS can be significantly enhanced by applying ideas from other discrete optimization methods such as branch-and-bound. Therefore, we believe that DPMS opens a promising research branch that awaits new ideas and improvements. 

\section{Related Work}
\subsection{Dynamic Programming and MaxSAT}
Dynamic programming (DP) is widely used in Boolean reasoning problems such as satisfiability checking \cite{DPBDD,bddsat}, Boolean synthesis \cite{bdddror}, and model counting \cite{DPCounting}.  
In practice, the DP-based model counter, ADDMC \cite{addmc} tied for the first place  in the weighted track of the 2020 Model Counting Competition \cite{MCC2020}. 
 ADDMC was further enhanced to DPMC \cite{dpmc}  by decoupling the planning phase from the execution phase. 
DPMC uses project-join trees to enclose various planning details such as variable ordering, making the planning phase a black box which can be applied across different Boolean optimization problems. Notwithstanding the success of DP-based model-counters, the practical potential of DP has not been deeply investigated in the MaxSAT community yet, though in \cite{anotherDPMaxSAT} the authors built a proof-of-concept DP-based MaxSAT solver and tested it on a limited range of instances.

\subsection{Existing Approaches for Solving General MaxSAT and How DPMS Compares}
First, hybrid constraints are admitted by DPMS via using ADDs to represent  pseudo-Boolean functions.
ADDs can compactly express many useful types of constraints besides disjunctive clauses, such as XOR and cardinality constraints. 
There exist other approaches for handling specific Max-hybrid-SAT problems; e.g., using stochastic local search in an incomplete Max-XOR-SAT solver \cite{XOR-SAT} and applying UNSAT-based approaches for Max-PB-SAT 
\cite{pboConstraint}. 
We are not aware, however, of the existence of a general solver for hybrid constraints. 
Alternatively, Max-hybrid-SAT can be reduced to group MaxSAT \cite{groupMaxSAT} by CNF encodings \cite{encoding-handbook-of-satisfiability}. Each hybrid constraint is first encoded to a group of disjunctive clauses. A new block variable is introduced for each group such that if the original hybrid constraint is violated, the corresponding group can only contribute one violated clause. Then, the group MaxSAT problem can be solved either as is, or further reduced to weighted MaxSAT. 
Those approaches, however, suffer from $i)$ significant increment of the problem size due to encodings, and $ii)$ the uncertain performance of choosing a specific encoding. 
In contrast, DPMS does not involve new variables, constraints, or encoding selection. 

Second, DPMS naturally handles Min-MaxSAT by using a graded project-join tree as the plan for execution, which is also applied in \cite{procount} for projected model counting. 
Graded project-join trees enforce the order of variable elimination restricted by the order of $\min$ and $\max$, while the execution phase remains the same as that on an ungraded tree. There exists a line of research that uses $d$-DNNF compilation with the \emph{constrained} property \cite{adnanDNNF}, which is similar to the graded tree in our work. Nevertheless, their approach only applies to CNF formulas, and the constrained property is not always enforced. 
In \cite{QMaxSAT,QMAXSAT2}, the authors studied the more general problem, Quantified MaxSAT, which falls to the complexity class PSPACE-complete. 
They did not exploit, however, the specific properties of Min-MaxSAT. 




\section{Notations and Preliminaries}
\label{section:pre}
\subsection{Pseudo-Boolean Functions}

\begin{definition} 
A pseudo-Boolean function $F$ over a variable set $X=\{x_1,\cdots,x_n\}$ is a mapping from the Boolean cube $\mathbb{B}^{X}$ to the real domain $\mathbb{R}$. 
\end{definition}

 We use $\mathbb{B}^X$ to denote both the power set of $X$ as well as the set of all assignments of a pseudo-Boolean function. I.e., an assignment $b\in \mathbb{B}^X$ is a subset of $X$. 
 Below we list several operations on pseudo-Boolean functions. 
\begin{definition} {\rm(Sum)} Let $F$, $G:\mathbb{B}^{X}\to \mathbb{R}$ be two pseudo-Boolean functions. The sum of $F$ and $G$, denoted by $F+G$, is defined by
$(F+G)({b})=F({b})+G({b})$ for all ${b}\in \mathbb{B}^{X}$.
\end{definition}

\begin{definition} {\rm(Optimization)}
Let $F$ be a pseudo-Boolean function, $x\in X$ be a variable.  
The maximization of $F$ w.r.t. $x$, 
denoted by $\mathop{\max}\limits_{x}F:\mathbb{B}^{X\setminus\{x\}}\to\mathbb{R}$, is defined by 

    $$(\mathop{\max}\limits_{x}F)({b})=\mathop{\max}\{F({b}\cup \{x\}),F({b})\}, \text{\,for all  ${b}\in \mathbb{B}^{X\setminus\{x\}}$.}$$
    
    The maximization operation can be extended to be w.r.t. a set $S\subseteq X$ of variables, denoted by
    $$\mathop{\max}_{S}F:\mathbb{B}^{X\setminus S}\to \mathbb{R}:
    (\mathop{\max}\limits_{S}F)({b}_1)=\mathop{\max}\limits_{{b}_2\in \mathbb{B}^{S}}F({b}_1 \cup {b}_2),\text{\,\, for all ${b}_1\in \mathbb{B}^{X\setminus S}$.}$$
      
The minimization of $F$ w.r.t. $x$ (resp.,  $S$) is defined similarly. After optimization, $x$ (resp., variables in $S$) is (are) \textbf{eliminated} from the domain of resulting function. 
We also define the $\argmax$  of $F$ w.r.t. $x$, denoted by $\argmax_xF: \mathbb{B}^{X\setminus \{x\}}\to \mathbb{B}^{\{x\}}$, for all  ${b}\in \mathbb{B}^{X\setminus\{x\}}$:

\begin{align*}
(\mathop{\argmax}\limits_xF)({b}) = 
\begin{cases}
\{x\} &\text{\,\,\,if $F({b}\cup \{x\})\ge F({b})$},\\
\emptyset &\text{\,\,\,otherwise.}
\end{cases}
\end{align*}
  \label{defn:optimization}
\end{definition}

 \begin{definition} {\rm (Derivative) }
Let $F$ be a pseudo-Boolean function and $x\in X$ be a variable. 
The derivative of $F$ w.r.t. $x$, denoted by $\nabla_{x}F:\mathbb{B}^{X\setminus \{x\}}\to \mathbb{R}$, is defined by
$
\nabla_{x}F({b}) =
F({b}\cup \{x\}) - F({b}), \text{\,\,for all ${b}\in \mathbb{B}^{X\setminus\{x\}}$.}
$  We say $x$ is  \emph{irrelevant} w.r.t. $F$, if $\nabla_{x}F\equiv0$.
\label{defn:derivative}
\end{definition}

 \begin{definition} {\rm (Sign) }
Let $F$ be a pseudo-Boolean function. 
The sign of $F$, denoted by $\texttt{sgn}F:\mathbb{B}^{X}\to \{0,1\}$, equals $1$ if $F({b})\ge 0$ and $0$ otherwise,
for all ${b}\in \mathbb{B}^{X}$. 
\label{defn:sign}
\end{definition}

\subsection{MaxSAT, Max-hybrid-SAT and Min-MaxSAT}
We consider the conjunctive form over a set of variables $X$, i.e., 
$f(X)=c_1(X)\wedge c_2(X)\wedge \dots \wedge c_m(X)$,
where each $c_i$ is a Boolean constraint. 
If each constraint of $f$ is a disjunctive clause, then $f$ is in the well-known CNF format. 
Traditional MaxSAT  asks the maximum number of constraints satisfied simultaneously by an assignment of $X$. 
Weighted MaxSAT attaches a real number  as the weight to each constraint  and seeks for the maximum total weight of satisfied constraints. 
 
Formally, MaxSAT can be written as an optimization problem on pseudo-Boolean functions. 
Let the constraint set of $f$ be $C_f$ and $w:C_f\to \mathbb{R}$ be a constraint weight function. 
For each formula in conjunctive form, we can construct an objective function as follows. 
For each $c\in C_f$ and assignment ${b}\in \mathbb{B}^X$, we overload  an indicator function to $c$: $c(b)=w(c)$ if $b$ satisfies $c$ and $0$ otherwise. Note that the constraint weight is included in the indicator function. 
Then the objective function is defined as follows. 

\begin{definition} {\rm (Objective function)}
Let $f$ be a conjunctive formula over $X$ with constraint set $C_f$. Then the objective function w.r.t. $f$, denoted by $F_f$ is defined as:

$$
F_f({b})= \sum_{c\in C_f}{c({b})}, \,\,\text{for all ${b}\in \mathbb{B}^X$.}
$$

\label{defn:objectiveFunction}
\end{definition}

\begin{definition} {\rm(MaxSAT)}
Using the notation in Definition \ref{defn:objectiveFunction}, the MaxSAT problem is to compute the value of:

$$
\mathop{\max}\limits_{X}F_f=
\mathop{\max}\limits_{{b}\in \mathbb{B}^{X}} \sum_{c\in C_f}{c({b})}.
$$

It is often desirable to also obtain a Boolean assignment, called a maximizer, ${b}^\star\in \mathbb{B}^X$ such that $F_f({b}^\star) = \mathop{\max}_{X}F_f$. 
\label{defn:maxsat}
\end{definition}

In this work, the constraints of a formula are not limited to clauses, yielding the \emph{Max-hybrid-SAT problem}. Specifically, each constraint can be of a type listed in Table  \ref{table:typesOfConstraints}.

	\begin{table}
		\centering
		\begin{footnotesize}
		\begin{tabular}{c c c c c}
			\toprule 
		Type & & Example  & & Size of ADD \\
			\cmidrule{1-1} \cmidrule{3-3} \cmidrule{5-5}
		CNF clause & & $x_1\vee\neg x_2 $ && $O(l)$\vspace{0.1cm}   \\ 
		XOR & & 	$x_1\oplus x_2\oplus x_3$ &&$O(l)$ \vspace{0.1cm} \\ 
			cardinality & &  \vspace{0.1cm} $\sum\limits_{i=1}^5x_i\ge 3$ & & $O(l^2)$ \\ 
			pseudo-Bool. & & $3x_1+5x_2-6x_3\ge 2$  && $O(M\cdot l)$\\
			\bottomrule
		\end{tabular} 
			\caption{ Types of  constraints considered in this work. In the third column, $l$ is \# variables in the constraint and $M$ is the largest magnitude of coefficients in the PB constraint.}
		\label{table:typesOfConstraints}
			\end{footnotesize}
	\end{table}

	\begin{table}[t!]
		\centering
			\begin{footnotesize}
		\begin{tabular}{c c c}
			\toprule 
			ADD Operation & & Complexity \\
			\cmidrule{1-1} \cmidrule{3-3}
			$\texttt{sum}(A_1,A_2)$ & & $O(|A_1|\cdot |A_2|)$ \vspace{0.05cm}   \\ 
		$\nabla_xA$ / $\max_xA$ / $\min_xA$ & & $O(|A|^2)$ \vspace{0.05cm} \\ 
		$\texttt{sgn}(A)$ & & $O(|A|\cdot \log(|A|))$  \\ 
			\bottomrule
		\end{tabular} 
			\caption{Complexity of some ADD operations}
		\label{table:complexityADDOperations}
			\end{footnotesize}
	\end{table}
	
By analogy with QBF \cite{QBF} in satisfiability checking, MaxSAT can be generalized by allowing both $\min$ and $\max$ operators. 

\begin{definition} {\rm(Min-MaxSAT)}
For $f$ in Boolean conjunctive form over $X\cup Y$, where $\{X,Y\}$ is a partition of all variables. The Min-MaxSAT problem is to compute:

$$
\mathop{\min}\limits_{X}\mathop{\max}\limits_{Y} F_f=\mathop{\min}\limits_{{b_1}\in\mathbb{B}^{X}}\mathop{\max}\limits_{{b_2}\in\mathbb{B}^{Y}} \sum_{c\in C_f}{c({b_1}\cup{b_2})}.
$$

The maximizer is a function $\mathcal{M}:\mathbb{B}^{X}\to \mathbb{B}^{Y}$ such that  $F_f({b_1}\cup \mathcal{M}({b_1}))=\mathop{\max}_{{b_2}\in \mathbb{B}^{Y}}F_f({b_1}\cup{b_2})$ for all ${b_1}\in \mathbb{B}^{X}$. 
\label{defn:minmax}
\end{definition}

The value $\mathop{\min}_{{b_1}\in\mathbb{B}^{X}}\mathop{\max}_{{b_2}\in\mathbb{B}^{Y}} F({b_1},{b_2})$ equals $\theta$ is interpreted as: for every assignment of $X$, there is always an assignment of $Y$ such that at least $\theta$ constraints are satisfied. Note that the MinSAT problem \cite{minsat}, which asks for an assignment that satisfies minimum number of constraints, is a special case of Min-MaxSAT.

When both $\min$ and $\max$ are applied, the order of operators becomes critical. Within a prefix, two adjacent operators  of different type ($\max$-$\min$ or $\min$-$\max$) is called an \emph{alternation} \cite{QMAXSAT2}.   MaxSAT has zero alternation, while a Min-MaxSAT instance has one. General cases  with more than one alternation  
is beyond the scope of this paper.
\subsection{Algebraic Decision Diagrams}  Algebraic Decision Diagram (ADD) \cite{algebraicDD} can be seen as  an extension of (reduced, ordered) \emph{Binary Decision Diagrams} (BDDs) \cite{BDD} to the real domain for representing pseudo-Boolean functions in a point-value style. An ADD is a directed acyclic graph, where each terminal node is associated with a real value. The size of an ADD is usually defined by the number of nodes. ADD supports multiple operations in polynomial time w.r.t. the ADD size \cite{BDDComplexity}, as Table \ref{table:complexityADDOperations}  shows.

\section{Theoretical Framework}
\subsection{Solving MaxSAT by DP and Project-Join Trees}
Our symbolic approach for solving the MaxSAT problem in Definition \ref{defn:maxsat} is based on applying two operations on pseudo-Boolean functions:
sum and optimization. In the general case, all optimization operations have to be conducted after all sum operations are completed. Nevertheless, sometimes it is possible and advantageous to apply optimization before sum. This is enabled by exploiting the structure of the conjunctive form and is called \emph{early optimization}. 

\begin{proposition} {\rm(Early Optimization)}
Let $F$, $G$: $\mathbb{B}^X\to \mathbb{R}$ be two pseudo-Boolean functions and $x\in X$ be a variable. If $\nabla_{x} G\equiv 0$, then we have
$
\mathop{\max}_{x} (F+G)\equiv (\mathop{\max}_{x}F)+G.
$
A similar result holds for the $\min$ operator. 
\label{prop:earlyMaximization}
\end{proposition}
In particular, if a variable $x$ only appears in a part of the additive objective function, then we only need to eliminate $x$ in components where $x$ is relevant.

Early elimination of variables, which reduces the problem size, has been proven successful and critical for other symbolic tasks.  
Since in MaxSAT, all variables are under the same operator ($\max$), theoretically the order of variable elimination can be arbitrary. In practice, however, the order of variable elimination is observed as crucial, potentially leading to exponential differences in running time \cite{variableOrder}. Therefore, an efficient algorithm needs to strategically choose the next variable to eliminate and numerous heuristics have been designed for finding a good order as a plan of execution \cite{ADDVariableOrder}. A recent work called DPMC decouples the planning phase from the execution phase for solving the model counting problem. In the planning phase, a \emph{project-join tree} is constructed as an execution plan. During the execution phase, constraints are joined using the product, and variables are eliminated according to the project-join tree. In this work, we reuse project-join trees for solving generalized MaxSAT problems. 
\begin{definition}  {\rm(Project-join tree) \cite{dpmc}} For a tree $T$, let $V(T)$ and $L(T)$ be the set of nodes and leaves of $T$, respectively. Let $f$ be a conjunctive formula over $X$. A project-join tree of $f$ is a tuple $\mathcal{T}=(T,r,\gamma,\pi)$, where $T$ is a tree with root $r\in V(T)$, $\gamma:L(T)\to C_f$ is a bijection between the leaves of $T$ and the  (indicator functions of) constraints of $f$, and $\pi:V(T)\setminus L(T)\to 2^{X}$ is a labeling function on internal nodes.
$\mathcal{T}$ must also satisfy
\begin{enumerate}    
\item  $\{\pi(v):v\in V(T)\setminus L(T)\}$ is a partition of $X$.
\item 
 For each internal node $v\in V(T)\setminus L(T)$, a variable $x\in \pi(v)$, and a constraint $c\in C_f$ that contains variable $x$, the leaf node corresponding to $c$, i.e., $\gamma^{-1}(c)$, must be a  descendant of $v$ in $T$.
\end{enumerate}

\end{definition}

The idea of using a project-join tree for solving MaxSAT is as follows. Each leaf of the tree corresponds to an original constraint, stored by $\gamma$. Each internal node $v$ corresponds to the pseudo-Boolean function obtained by   summing (join) all sub-functions from the children of $v$ and eliminating all variables attached to the node $v$ by maximization (project). This idea is formalized as the \emph{valuation}.

\begin{definition}  {\rm(Valuation)} For a project-join tree $\mathcal{T}=(T,r,\gamma,\pi)$, a node $v\in V(T)$, the valuation of $v$, which is a pseudo-Boolean function denoted by $\mathcal{F}_v^{\mathcal{T}}$, is defined as:

\begin{align}\nonumber
\mathcal{F}_v^\mathcal{T}=
    \begin{cases}
           \gamma(v)  &\text{if $v\in L(T)$},\\
           \mathop{\max}\limits_{\pi(v)}\sum\limits_{u\in \texttt{children}(v)}\mathcal{F}_u^\mathcal{T} &\text{if $v\not\in L(T)$}.
    \end{cases}
\end{align}

\label{defn:valuation}
\end{definition}

The valuation of the root $r$ gives the answer to the MaxSAT problem. 
\begin{theorem} 
Let $\mathcal{T}$ be a project-join tree, then we have

$$
\mathcal{F}_r^{\mathcal{T}}=\mathop{\max}\limits_{X}F_f =\mathop{\max}\limits_{{b}\in \mathbb{B}^{X}} \sum_{c\in C_f}{c({b})}
$$

\label{theo:correctness}
\end{theorem}

A project-join tree, which can be viewed as an encapsulation of many planning details, e.g., cluster ordering, 
guides the execution of a DP-based approach \cite{dpmc}.
Since building a project-join tree only depends on the incidence graph \cite{incidenceGraph} of a formula, we use the tree constructor from DPMC as a black box for solving MaxSAT. 

\subsection{Using ADDs to Express Pseudo-Boolean Functions}
Solving Max-hybrid-SAT symbolically as discussed above requires the data structure for pseudo-Boolean functions to have $i$) the support of efficient functional addition and maximization; $ii$) the expressiveness of different types of  constraints. ADD is a promising candidate, compared, for example, with  truth table and polynomial representation such as Fourier expansions \cite{O'Donnell:2014:ABF:2683783}. In terms of size, ADDs are usually more compact than other representations due to redundancy reduction. In particular, ADDs can succinctly encode a number of useful types of constraints (some examples are listed in Table \ref{table:typesOfConstraints} \cite{fouriersat,GradSAT}). Moreover, ADDs support several important operations efficiently, as Table \ref{table:complexityADDOperations} shows. ADDs also own well-developed packages such as CUDD \cite{cudd} and Sylvan \cite{sylvan}.   

Summing up all ADDs corresponding to initial constraints in $C_f$--without early variable eliminations--already provides a complete algorithm for MaxSAT. To see this, note that after summing all ADDs obtained from $C_f$, the terminal nodes of the monolithic ADD represent the cost of all assignments. Therefore, one just needs to find the terminal value of the monolithic ADD with the largest value by maximizing over this ADD.   Using the sum operation first, however, can lead to a potential exponential blow-up of the ADD size. 
Thus, early optimization introduced above has to be involved in order to alleviate the explosion.

 At each internal node $v$ of the project-join tree, we need to compute the corresponding valuation $\mathcal{F}_v^{\mathcal{T}}$.
We first sum up sub-valuations from the children of $v$ to get an ADD $A$. Then we iteratively \emph{apply}  $\max$ w.r.t. variables attached on $v$, i.e., $\pi(v)$, to $A$. This method naturally follows  Definition \ref{defn:valuation} and is shown in Algorithm \ref{algo:computeValuationByADDMax}. Line \ref{line:algo2line5}  and \ref{line:algo2line6} are for constructing the maximizer, with details provided later.

\begin{algorithm}[ht]
    \SetAlgoLined
    \SetKwInOut{Input}{Input}
    \SetKwInOut{Output}{Output}
    \Input{A conjunctive formula $f$ with set of constraints $C_f$}
    \Output{The value of $\max_XF_f$ and a maximizer ${b}^\star$}
    \vspace{0.1cm}
    \hrule
    \vspace{0.1cm}
    $\mathcal{T}=(T,r,\gamma,\pi)\gets $\texttt{constructProjectJoinTree}$(C_f)$\\
     $\mathcal{S}\gets$  an empty stack\\
    $answer\gets\texttt{computeValuation}(\mathcal{T},r,\mathcal{S})$\\
    $b^\star\gets\texttt{constructMaximizer}(\mathcal{S})$\\
    \Return  $answer, {b}^\star$
 \caption{\texttt{DPMS}$(f)$}
 \label{algorithm:algorithm:ADD-Based-MaxSAT-Solving}
\end{algorithm}

\begin{algorithm}[ht]
    \SetAlgoLined
    \SetKwInOut{Input}{Input}
    \SetKwInOut{Output}{Output}
    \Input{A project-join tree $\mathcal{T}=(T,r,\gamma,\pi)$, a node $v\in V(T)$, a stack $\mathcal{S}$ for constructing the maximizer.}
    \Output{An ADD representation of  $\mathcal{F}^\mathcal{T}_v$}
    \vspace{0.1cm}
    \hrule
    \vspace{0.1cm}
   \lIf{$v\in L(T)$ } 
    {\Return $\texttt{compileADD}(\gamma(v))$\label{line:algo2line1}} 
    $A\gets \sum\limits_{u\in \texttt{children(v)}}\texttt{computeValuation}(\mathcal{T},u,\mathcal{S})$ \label{line:algo2line2}\\
      \For{$x\in \pi(v)$}{    
      \text{ \texttt{/* For maximizer generation */} }\\
       $G_x\gets \texttt{sgn}(\nabla_xA)$ \label{line:algo2line5}\\
         $\mathcal{S}$.\texttt{push}$((x,G_x))$\label{line:algo2line6}\\
    $A\gets \mathop{\max}_{x}A$     \text{ \texttt{// Eliminate $x$} }\\ 
    }
  \Return $A$ \\
 \caption{\texttt{computeValuation}$(\mathcal{T},v,\mathcal{S})$ }
 \label{algo:computeValuationByADDMax}
\end{algorithm}

Inspired by \emph{Basic Algorithm} of pseudo-Boolean optimization \cite{basicalgorithm2}, we consider an alternative of computing the same output with Alg. \ref{algo:computeValuationByADDMax}. The theory of Basic Algorithm and practical comparisons between Alg. \ref{algo:computeValuationByADDMax} and the alternative can be found in the appendix. 

\subsection{Solving Min-MaxSAT by Graded Trees}
In MaxSAT, only the $\max$ operator is used. Nevertheless, since it is equally expensive to compute $\max$ and $\min$ for the symbolic approach, adapting techniques described above for solving the Min-MaxSAT problem in Definition \ref{defn:minmax} is desirable. In MaxSAT, the order of eliminating variables can be arbitrary, while that in Min-MaxSAT is restricted. More specifically, a variable under $\min$ can be eliminated only after all variables under $\max$ are eliminated. In terms of the project-join tree, roughly speaking, in every branch, a node attached with $\min$ variables must be ``higher" than all nodes attached with $\max$ variables. This requirement can be satisfied by the \emph{graded} project-join tree, originally used for \emph{projected model counting}~\cite{procount}.

\begin{definition} {\rm (Graded project-join tree) \cite{procount}} Let $X,Y$ be a partition of all variables. A project-join tree $\mathcal{T}=(T,r,\gamma,\pi)$ is called ($X$-$Y$) graded if there exists $I_X$, $I_Y\subseteq V(T)\setminus L(T)$, such that:
\begin{enumerate}
    \item 
 $I_X,I_Y$ is a partition of internal nodes $V(T)\setminus L(T)$.
    \item 
 For each internal node $v\in V(T)\setminus L(T)$, if $v\in I_{X}$, then $\pi(v)\subseteq X$; if $v\in I_{Y}$ then $\pi(v)\subseteq Y$.
    \item 
  For every pair of internal nodes $(u,v)$ such that $u\in I_x$ and $v\in I_Y$, $u$ is not a  descendant of $v$ in $T$.
\end{enumerate}
\end{definition}
Constructing a graded project-join tree also only depends on the incidence graph of a formula. Thus we can apply existing graded project-join tree builders to our Min-MaxSAT problem, where $X$ is the set of $\min$ variables and $Y$  is the set of $\max$ variables~\cite{procount}. 

We note that DPMS can easily handle also Max-MinSAT problems by making $X$ the set of $max$ variables and $Y$ the set of $min$ variables. In contrast, adapting state-of-the-art MaxSAT and PB solvers to solve Max-MinSAT can be nontrivial.

\subsection{Constructing the Maximizer}
MaxSAT often asks also for the maximizer ${b}^\star$ such that $F_f({b}^\star)=\mathop{\max}_XF_f$
\cite{maxsatHandbook}. In DPMS, the maximizer is constructed recursively in the reverse order of variable elimination, based on the following proposition.
\begin{proposition}
For a pseudo-Boolean function $F$ and a variable $x\in X$, if $b^\star$ is a maximizer of $\max_xF$, then $\{(\argmax_xF)(b^\star)\}\cup b^\star$ is a maximizer of $F$.
\label{prop:maximizer}
\end{proposition}
In the following proposition, we show that the $\argmax$ of a pseudo-Boolean function is the \texttt{sgn} of the derivative.
\begin{proposition} For a pseudo-Boolean function $F:\mathbb{B}^X\to \mathbb{R}$ and a variable $x$, we have, for all ${b}\in \mathbb{B}^{X\setminus\{x\}}$: $\mathop{\argmax}_xF({b})$ equals $\{x\}$ if $\texttt{sgn}\nabla_xF({b})= 1$ and $\emptyset$ otherwise.
\label{prop:argmaxAndDerivative}
\end{proposition}
The idea of generating a maximizer in DPMS by Algorithm \ref{algo:computeValuationByADDMax} and \ref{algo:maximizer} is as follows. In Algorithm \ref{algo:computeValuationByADDMax}, suppose when a variable $x$ is eliminated, variables in $S\subseteq X$ have already been summed-out. Then $\nabla_xA$ is equivalent to $\nabla_x\max_SF$ (see the proof of Theorem \ref{theo:maximizerAlg}). Therefore,  by Proposition \ref{prop:argmaxAndDerivative},
the $0$-$1$ ADD $G_x=\texttt{sgn}(\nabla_xA)$ computed in Line \ref{line:algo2line5}, is the ADD representation of $\argmax_x(\max_SF)$.  By Proposition \ref{prop:maximizer}, with $G_x$,
  finding a maximizer of $max_SF$ can be reduced to that of  $(max_{S\cup\{x\}}F)$, a smaller problem.  As variable-ADD pairs get pushed into the stack in Line \ref{line:algo2line6}, the ADD in the pair relies on fewer and fewer variables. In particular, when the last variable, say $x_{l}$ is eliminated, then  $G_{x_{l}}$ is a constant $0$-$1$ ADD indicating the value of $x_{l}$ in the maximizer. The stack $\mathcal{S}$ will pop variables in the reverse order of variable elimination.  In Algorithm  \ref{algo:maximizer}, a maximizer is iteratively built based on Proposition \ref{prop:maximizer}. Similar idea of building a maximizer  can be found in  \emph{Basic Algorithm} of pseudo-Boolean optimization \cite{basicalgorithm2}.
\begin{algorithm}[]
    \SetAlgoLined
    \SetKwInOut{Input}{Input}
    \SetKwInOut{Output}{Output}
    \Input{A stack $\mathcal{S}$ of $|X|$ variable-ADD pairs}
    \Output{${b}^\star$ such that $F_f({b}^\star)=\mathop{\max}_XF_f$}
    \vspace{0.1cm}
    \hrule
    \vspace{0.1cm}
    $b^\star \gets \emptyset$\\
    \While{$\mathcal{S}$ is not empty}{
    $(x,G_x) \gets \mathcal{S}.\texttt{pop}()$\\
    \lIf{$G_x(b^\star)=1$}{
         $b^\star\gets b^\star \cup \{x\}$\label{line:algo3line4}
    }
    }
    \Return  ${b}^\star$\\
 \caption{$\texttt{constructMaximizer}(\mathcal{S})$}
 \label{algo:maximizer}
\end{algorithm}
\begin{theorem}
Algorithm \ref{algo:maximizer} correctly  returns a maximizer.
\label{theo:maximizerAlg}
\end{theorem}
In Min-MaxSAT, the maximizer is a function from variables under $\min$ to those under $\max$, which is similar to Skolem functions in 2-QBF \cite{2qbf}. Synthesizing such a  maximizer could be non-trivial for non-symbolic methods. In DPMS, this can be done by Algorithm \ref{algo:maximizer} but only $\max$ variables are pushed into the stack $\mathcal{S}$.

\subsection{The Complexity of DPMS}
 For a (graded) project-join tree $\mathcal{T}=(T,r,\gamma,\pi)$, we define the \emph{size} of a node $v$, $size(v)$, to be \# variables in the valuation $\mathcal{F}_v^{\mathcal{T}}$, plus \# variables eliminated in $v$. The width of a tree $\mathcal{T}$, $width(\mathcal{T})$, is defined to be $\max_{v\in V(T)} size(v)$. Roughly speaking, $width(\mathcal{T})$ is the maximum \# variables in a single ADD during the execution.  The width is an upper bound of the \emph{treewidth} of the incidence graph of the formula.
It is well known that the running time of DP-based approaches can be bounded by a polynomial over the input size on problems with constant width \cite{rankwidth}.
\begin{theorem}
If the project-join tree (resp. graded project-join tree) builder returns a tree with width $k$, then DPMS  solves the MaxSAT (resp. Min-MaxSAT) problem in time $O(k\cdot 2^{k}\cdot|f|)$, where $|f|$ is the size of the conjunctive formula.
\label{theo:treewidth}
\end{theorem}

Hence, DPMS exhibits the potential of exploiting the low-width property of instances, as shown later in Section \ref{sec:exp}. 

\subsection{An Optimization of DPMS: Pruning ADDs by Pre-Computed Bounds}
\label{subsection:pruning}
The main barrier to the efficiency of DPMS is the explosive size of the intermediate ADDs. We alleviate this issue by pruning ADDs via pre-computed bounds of the optimal cost. The idea is, if we know that a partial assignment can not be extended to a maximizer, then it is safe to prune the corresponding branch from all intermediate ADDs. This idea can be viewed as applying branch-and-bound \cite{branchAndBound}
to ADDs.  For a MaxSAT instance, suppose we obtain an upper bound of the cost of the maximizer, say $U$. Then  for each individual ADD $A$, if the value of a terminal node $v$ indicates that the cost of the corresponding partial assignment in $A$ already exceeds $U$, then it is impossible to extend this partial assignment to a maximizer. In this case, we  set the value of $v$ to $-\infty$, which provides two advantages. First,  the number of terminal values in $A$ is reduced, which usually decreases the ADD size. Second, the partial assignment w.r.t. $v$ is ``blocked" globally by summing $A$ with other ADDs in the rest execution of the algorithm. In the experiment section, we show that using a relatively good bound significantly enhances DPMS. 

One can obtain $U$  in several ways. One way is to run an incomplete local search solver within a time limit. Another way is to read $U$ directly from the problem instance, e.g., the total weight of soft constraints of a partial MaxSAT instance.

\section{Empirical Results}
\label{sec:exp}
We aim to answer the following research questions:

\noindent\textbf{RQ1.} Can DPMS outperform state-of-the-art MaxSAT and pseudo-Boolean solvers on certain problems?

\noindent\textbf{RQ2.} How do pre-computed bounds and ADD pruning improve the performance of DPMS?

\noindent\textbf{RQ3.} How does DPMS perform on real-life benchmarks?

To answer the three research questions above, we designed three experiments, respectively. The first two experiments used synthetic and random hybrid formulas, respectively. The third experiment gathered instances from Boolean reasoning competitions.

We included 10 state-of-the-art MaxSAT solvers and 3 pseudo-Boolean (PB) solvers in comparison, which cover a wide range of techniques such as SAT-based, MIP, and B\&B. We list all the solver competitors below.

\nonumber\paragraph{10 MaxSAT solvers}
\begin{itemize}
    \item 7 Competitors from MaxSAT Evaluation 2021: MaxHS \cite{maxhs}, CashWMaxSAT, EvalMaxSAT \cite{evalmaxsat}, OpenWBO \cite{openwbo}, UwrMaxSAT \cite{uwrmaxsat}, Exact, and Pacose \cite{pacose}.
    \item MaxCDCL \cite{maxcdcl}, a recent MaxSAT solver based on  B\&B and CDCL.
    \item ILPMaxSAT \cite{ILPMaxSAT}, an Integer-Linear-Programming-based MaxSAT solver.
    \item GaussMaxHS \cite{gaussmaxhs}, a CNF-XOR MaxSAT solver which accepts hard XOR  constraints. 
\end{itemize}

\paragraph{3 Pseudo-Boolean solvers}
\begin{itemize}
    \item NAPS \cite{naps}, winner of the WBO Track of PB competition 16.
    \item WBO \cite{openwbo}, a PB solver that accepts .wbo and .pbo format.
    \item CuttingCore~\cite{cuttingToTheCore}, a recent PB solver based on core-guided search and cutting planes reasoning.
\end{itemize}

\paragraph{1 QBF solver}
\begin{itemize}
    \item DepQBF \cite{depqbf}, medal winner of QBF evaluations.
\end{itemize}

Since MaxSAT and PB solvers used in comparison can not handle  Max-hybrid-SAT format, we reduce those hybrid instances to \emph{group MaxSAT} according to \cite{groupMaxSAT}. Then, in order to compare DPMS with MaxSAT and PB solvers, group MaxSAT problems were further reduced to weighted MaxSAT and pseudo-Boolean optimization problems by adding \emph{block variables}.   

All experiments were run on single CPU cores of a Linux cluster at 2.60-GHz and with 16 GB of RAM. We implemented DPMS borrowing from the DPMC \cite{dpmc} and ProCount \cite{procount} code base. DPMC's default setting  was used as the default settings for DPMS. FlowCutter~\cite{flowCutter} was used as the tree builder. Time limit was set to be 400 seconds unless specified. 

\textbf{Experiment 1: Evaluation on  low-width Chain Max-Hybrid-SAT instances.} We generated the following unweighted ``chain'' formulas with certain width. A formula with number of variables $n$ and width parameter $k$ contains $(n-k+1)$ constraints, where the $i$-th constraint exactly contains variables in $\{x_i,x_{i+1},\cdots,x_{i+k-1}\}$. Every constraint is  either an XOR with random polarity or a cardinality constraint with random right-hand-side, each with probability 0.5. It is easy to show that the width of such formulas is  $k$.  For $n\in\{100,200,400,600,800,1000,1200\}$, $k\in\{5,10,15,20,25,30\}$, we generated $20$ chain formulas with $n$ variables and width $k$. Total number of instances: 840.

\begin{figure}[ht!]
\centering
\includegraphics[width=0.6\columnwidth]{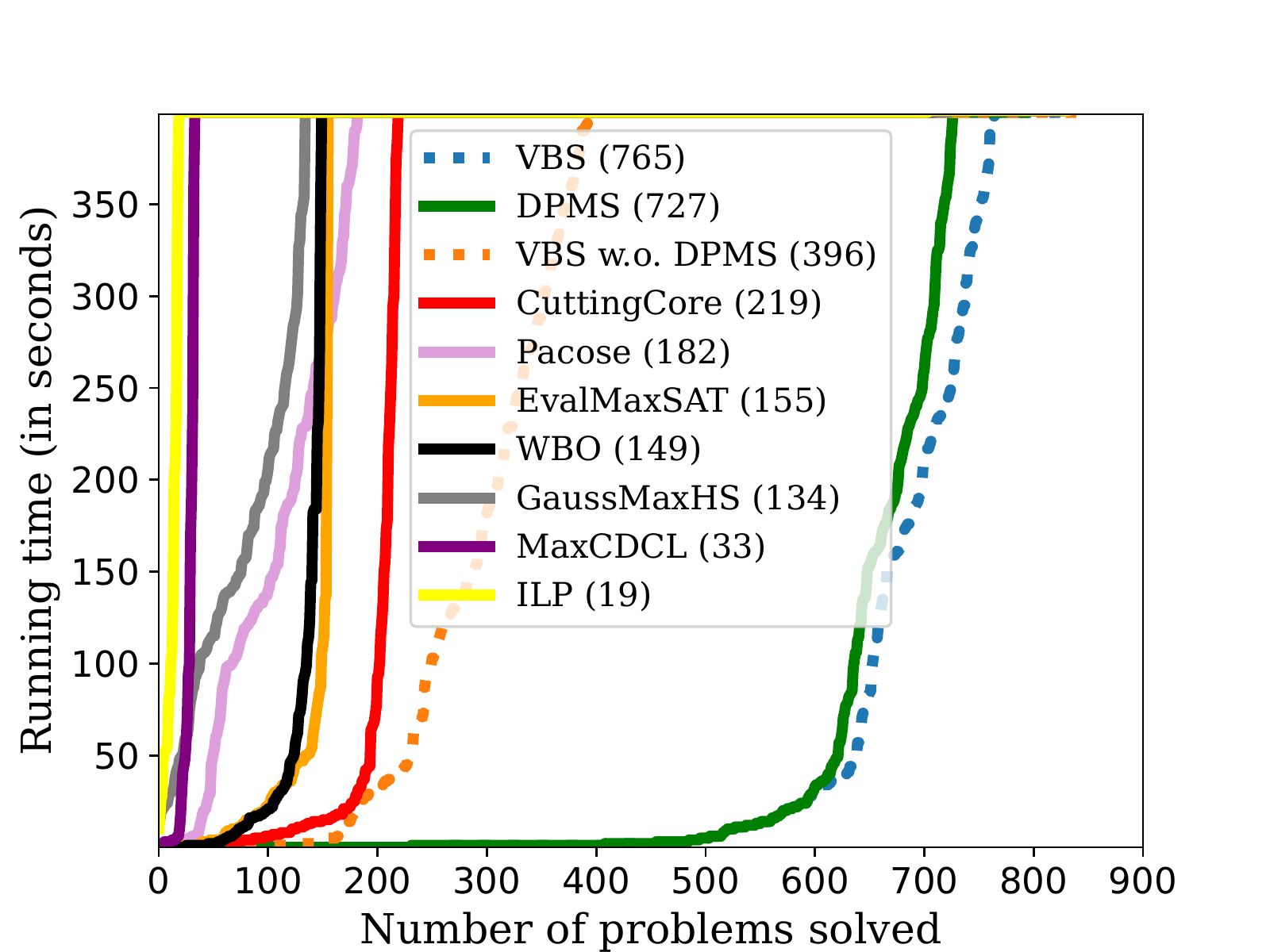}  
\caption{\textbf{Results on 840 hybrid chain formulas describied in Experiment 1.} (Best viewed online) The results of some solvers are presented in the figure.  \# solved instances of other solvers are: Exact (169),  Uwrmaxsat (163), OpenWBO (162),  MaxHS (75) and CashwMaxSAT (8). DPMS outperformed all solvers in comparison and significantly improved the VBS.}
\label{fig:chain}
\end{figure}

\textbf{Answer to RQ1.} The overall results of Experiment 1 are shown in Figure \ref{fig:chain}, where the \emph{virtual best solver} (VBS) and VBS without DPMS are also plotted. On the low-width hybrid MaxSAT instances, DPMS solved 727 out of 840 instances,  significantly more than all other MaxSAT and PB solvers in comparison and  the VBS (396) of them.  The results indicate that DPMS exhibits promising performance on hybrid low-width instances. Meanwhile, state-of-the-art solvers fail to leverage the low-width property and are considerably slowed down by explosive CNF/PB encodings. 

\textbf{Experiment 2: Evaluation on General Random Max-Hybrid-SAT instances.} In this experiment, we generated random  CARD-XOR and PB-XOR instances. For  $n\in\{50,75,100\}$, $\alpha_C,\alpha_X\in\{0.3,0.4,\cdots,1\}$, $k\in\{5,7\}$,  we generated $20$ instances with $n$ variables, $(\alpha_C\cdot n)$ random $k$-cardinality (resp. PB) constraints and $(\alpha_X \cdot n)$ random $k$-XOR  constraints. Totally number of instances: 15360.

Generally, the width of pure random instances grows as the number of variables increases. Previous research indicates that high-width problems can be hard for DP-based algorithms \cite{dpmc}. Therefore, we do not expect DPMS to outperform its competitors on the full set of benchmarks. Instead, we aim to show how ADD pruning can greatly enhance the performance of DPMS. SATLike \cite{satlike}, a pure local search MaxSAT solver, was used to provide an upper bound of cost. For each instance, we ran SATLike for 10 seconds (included in the total running clock time) and used the best bound found for ADD pruning. This combination of DPMS and SATLike is called DPMS + LS. 

	\begin{table}
		\centering
		\begin{footnotesize}
			\scalebox{1}{
		\begin{tabular}{c c c c c}
			\toprule 
		 & & DPMS  & & DPMS + LS \\
			\cmidrule{1-1} \cmidrule{3-3} \cmidrule{5-5}
		\# Solved instances & & 581 && 4385 \vspace{0.1cm}   \\ 
			\# Avg. width of solved instances & &  \vspace{0.1cm} 29.1 & & 38.2 \\
			 Avg. reduction of largest ADD size ($\times$) & &  \vspace{0.1cm} 1 & & 43.6 \\
			 Avg. Speed-up ($\times$) & &  \vspace{0.1cm} 1 & & 3.86 \\	   
			\bottomrule
		\end{tabular} 
		}
			\caption{Comparison between DPMS and DPMS + LS on random benchmarks. The average width of all instances is 45.7. The average approximation ratio given by the pure local search solver SATLike is 1.21. ADD pruning makes DPMS+LS solve considerably more instances than DPMS and increase the width DPMS can handle. }
		\label{table:exp2}
			\end{footnotesize}
	\end{table}
	
\textbf{Answer to RQ2.} The comparison between DPMS and  DPMS + LS is shown in Table \ref{table:exp2}. The average approximation ratio of the upper bound provided by running SATLike for 10 seconds is 1.21. With the help of SATLike and ADD pruning, DPMS + LS solved significantly more instances (4385) than DPMS (581). DPMS + LS can also handle instances with higher width (38.2) compared with DPMS (29.1). Moreover, DPMS + LS provides notable acceleration ($3.86\times$) and reduction of space usage ($43.6\times$). Therefore, we conclude that ADD pruning is an effective optimization of DPMS.

\textbf{Experiment 3: Min-MaxSAT, MaxSAT-PB-SAT, and MaxSAT instances from competitions.} 
 In this experiment, we aim to check whether DPMS can help solve real-life  problems. 
 
 To our best knowledge, there are neither Min-MaxSAT benchmarks nor implementations of Min-MaxSAT solvers publicly available. Thus, we gathered 853 instances from 2-QBF ($\forall\exists$) Track of recent QBF Evaluations (10, 16, 17, 18) and compared DPMS with DepQBF, a state-of-the-art QBF solver, though Min-MaxSAT is harder than QBF-SAT. 

 We also run DPMS on PB-WBO and MaxSAT instances from recent competitions. The PB-WBO instances are from the latest pseudo-Boolean Competition (PB’16), Weighted Boolean Optimization (WBO) Track, with the time limit set to 1800 seconds. Those instances can be viewed as weighted partial Max-PB-SAT problems.  The MaxSAT instances are from MaxSAT evaluation 2021 (MSE'21), Completed Weighted Track with the time limit set to 300 seconds. Most instances of those competitions are considerably large-size. 
 
\textbf{Answer to RQ3.} The results on Min-MaxSAT instances are listed in Table \ref{table:category2}. On all 853 2-QBF instances, the tree builder succeeded on 125 instances and DPMS solved 90 of them. All successfully built graded project-join trees have width smaller than 100. 
DPMS is faster than DepQBF on 48 instances (Row 6) with $15.7\times$ acceleration (Row 8). Furthermore, there are in total 39 instances uniquely solved by DPMS in the evaluation (Row 9). Moreover, all these 39 instances were solved in less than 10\% of the time limit while DepQBF timed out. 
 
The results of MaxSAT and PB evaluations are shown in Table \ref{table:PBMaxSAT}. DPMS with bounds given by local search solved 94 instances (623 in total) for MaxSAT instances. On Max-PB-SAT instances, DPMS with trivial bounds solved 97 instances (1059 in total). Instances solved by DPMS have lower average treewidth ($<40$) and a relatively short running time ($7$s for Max-PB-SAT and $23.6$s for MaxSAT).  DPMS is in general, however, not yet competitive on real-life MaxSAT and PB instances when compared with solvers from competitions. 

	\begin{table}
	\centering
		\begin{footnotesize}
		\scalebox{1}{
		\begin{tabular}{c c c}
			\toprule 
  Row &\# Instances    &   \\\cmidrule{1-3}
1 &	Total &   853   \vspace{0.05cm}\\  	\cmidrule{1-3}
2 &		Tree builder succeed  &   125   \vspace{0.05cm}\\ 
3 &		Avg. width of Row 2 & 58.8  \vspace{0.05cm}\\ 	\cmidrule{1-3}
4 &		Solved   by DPMS        & 90   \vspace{0.05cm} \\ 
5 &		Avg. width of Row 4 & 47.8 \vspace{0.05cm}\\  	\cmidrule{1-3}
6 &		DPMS is faster than DepQBF & 48  \vspace{0.05cm}\\
7 &		Avg. width of Row 6     & 62.6\vspace{0.05cm}\\ 
8 &		Avg. acce. ($\times$) of Row 6 & 15.7  \vspace{0.05cm}\\ 	\cmidrule{1-3}
9 &			 Solved Uniquely by DPMS   & 39    \vspace{0.05cm}\\ 
10 &		Avg. width of Row 9   & 67.1   \vspace{0.05cm}\\ 
11 &		Avg. time (s) of Row 9 & 4.13  \vspace{0.05cm}\\ 
		\bottomrule
		\end{tabular} }
		\caption{\textbf{Performance of DPMS on Min-MaxSAT instances, compared with DepQBF.} DPMS was capable of solving 39 instances where DepQBF timed out.}
			\label{table:category2}
		\end{footnotesize}
	\end{table}

 	\begin{table}
	\centering
		\begin{footnotesize}
		\scalebox{1}{
		\begin{tabular}{c c c c c c c}
			\toprule 
  Row & \#instances &  PB-WBO (DPMS+Trivial)    &  & MaxSAT (DPMS+LS)   \\\cmidrule{1-5}
1 &	Total &   1059  && 623 \vspace{0.05cm}\\  	
2 &		Tree builder succeed  &   181 & &116 \vspace{0.05cm}\\ 
3 &		Avg. width of Row 2 & 44.8 &  &43.2 \vspace{0.05cm}\\ 
4 &		Solved   by DPMS        & 97 & & 94  \vspace{0.05cm} \\ 
5 &		Avg. width of Row 4 & 28.4 &  & 38.2 \vspace{0.05cm}\\  	
6 &		Avg. running time (s) of Row 4 & 7.0   && 23.6 \vspace{0.05cm}\\ 	
\bottomrule
		\end{tabular} }
		\caption{\textbf{Performance of DPMS on PB-WBO and MaxSAT instances from competitions.} Instances solved by DPMS on average are low-width. The average running time of solved instances is also relatively low.}
			\label{table:PBMaxSAT}
		\end{footnotesize}
	\end{table}
 
\noindent\textbf{Summary.} We demonstrated that DPMS can beat a comprehensive set of state-of-the-art solvers on synthetic hybrid low-width instances. ADD pruning greatly enhances the performance of DPMS.  On Min-MaxSAT problems, DPMS captures the structural properties of some instances which are difficult for other algorithms to exploit. DPMS is not yet competitive against specialized solvers on real-life MaxSAT and PB instances. Overall, the versatility of handling various formulations as well as encouraging performance on certain benchmarks make DPMS a promising framework.

\section{Conclusion and Future Directions}
In this work, we proposed a novel and versatile ADD-based framework called DPMS for natively handling a number of generalized MaxSAT problems such as Max-hybrid-SAT and Min-MaxSAT. We leverage the similarity between max-of-sum and sum-of-product to apply the project-join-tree-based approaches in MaxSAT solving. ADD enables hybrid constraints, while graded trees enable  Min-Max formulations. Theoretical analysis indicates that DPMS scales polynomially on instances with bounded width. We implemented our method based on the DPMC code base. Empirical results demonstrate that DPMS outperforms state-of-the-art MaxSAT and PB solvers on synthetic hybrid MaxSAT instances with low width. Experiments also showed that DPMS can be significantly accelerated by branch-and-bound and ADD pruning. We believe DPMS opens a  new research branch and leads to many interesting future directions. For example, handling instances with alternation larger than $2$, i.e., general Quantified-MaxSAT \cite{QMaxSAT}, parallelizing DPMS by alternative tree decomposition tools and DD packages such as Sylvan \cite{sylvan}, and integrating other optimization techniques into this dynamic programming framework.

\newpage
\bibliography{ijcai22} 
\newpage
\appendix
\begin{appendix}
\section{The Basic Algorithm and an Alternative of Algorithm \ref{algo:computeValuationByADDMax}}
In this section, we show an alternative algorithm based on \emph{Basic Algorithm} that computes the same output as Algorithm \ref{algo:computeValuationByADDMax}. We first introduce Basic Algorithm of pseudo-Boolean optimization, whose idea is shown in Theorem \ref{theo:basicAlgorithm}.

\begin{theorem}{\rm (Maximize-out a variable by substitution) \cite{basicalgorithm1,basicalgorithm2}} For a pseudo-Boolean function $F:\mathbb{B}^X\to \mathbb{R}$ and a variable $x$. Then we have, for all ${b}\in \mathbb{B}^{X\setminus\{x\}}$:
\begin{equation}\nonumber
    \begin{split}
\mathop{\max}\limits_{x} F({b})= F({b}\cup \argmax_xF({b})))
 \end{split}
\end{equation}
\label{theo:basicAlgorithm}
Note that $F({b}\cup \argmax_xF({b})))$ can be viewed as substituting $x$ by $\argmax_xF({b})$  in $F$.
\end{theorem}

Theorem \ref{theo:basicAlgorithm} looks similar with Proposition \ref{prop:maximizer}. In fact, Theorem \ref{theo:basicAlgorithm} is a generalization of Proposition \ref{prop:maximizer} to all assignments rather than just the maximizer. One interesting property of Basic Algorithm in our case is, a variable can be eliminated only by substitution without the sum operation \footnote{Though the derivatives still need to be summed, as we will see later. The derivatives are, however, with smaller size than the original ADD.}. Therefore, substituting $x$ by $\argmax_xF$ can be done separately in each individual ADDs. The pseudo-Boolean $\argmax_xF$ is in fact computed by $G_x$ in the algorithm (see Lemma \ref{lemma:Gx} and its proof in below). Based on the above, we show a variant of Algorithm \ref{algo:computeValuationByADDMax}, namely Algorithm \ref{algo:computeValuationByBA} below. Note that the valuation of each node in Algorithm \ref{algo:computeValuationByBA} is no longer an ADD as in Algorithm \ref{algo:computeValuationByADDMax}, but a set of ADDs, whose sum is equivalent with the resulting ADD in Algorithm \ref{algo:computeValuationByADDMax}. 


Algorithm \ref{algo:computeValuationByBA} works as follows. By Proposition \ref{prop:argmaxAndDerivative}, the $\argmax$ operator can be implemented by the \texttt{sgn} of derivative, i.e., $G_x$ in the algorithm. Since $F_f$ is an additive objective function, its derivative $\nabla_xF$ can be computed by first obtaining the derivatives for all additive components and then summing up these derivatives (the $G_x$ in Algorithm \ref{algo:computeValuationByADDMax} and \ref{algo:computeValuationByBA} are equivalent). Afterwards, Algorithm \ref{algo:computeValuationByBA} substitutes $x$ by $G_x$ in all ADDs of sub-valuations (\texttt{Compose}$(A,x,G_x)$) and returns the set $A_v$. 

\begin{algorithm}[ht]
      \SetAlgoLined
    \SetKwInOut{Input}{Input}
    \SetKwInOut{Output}{Output}
     \Input{A project-join tree $\mathcal{T}=(T,r,\gamma,\pi)$, a node $v\in V(T)$, a stack $\mathcal{S}$ for constructing the maximizer.}
    \Output{A set of ADDs, whose sum is equivalent to  $\mathcal{F}^\mathcal{T}_v$}
    \vspace{0.1cm}
    \hrule
    \vspace{0.1cm}
   \lIf{$v\in L(T)$ } 
    {\Return \{ \texttt{compileADD}$(\gamma(v))\}$ \,\,\,\,// $v$ is a leaf}\label{line:algo3line1}
     \For{$u\in \texttt{children}(v)$}{
         $A_u\gets \texttt{computeValuation}(\mathcal{T},u,\mathcal{S})$\\
     }
      \For{$x\in \pi(v)$}{
       $D\gets \texttt{ZeroADD}$\\
          \For{$u\in \texttt{children}(v)$}{
             \For{$A\in A_u$}{
                  $D\gets D+\nabla_xA$\\
        }
    }
   $G_x\gets \texttt{sgn}(D)$\\
       $\mathcal{S}.\texttt{push}((x,G_x))$\\
     \For{$u\in \texttt{children}(v)$}{
        \For{$A\in A_u$}{
          $A\gets \texttt{Compose}(A,x,G_x)$\\
          }
        }
    }
    \Return $\bigcup_{u\in \texttt{children(v)}}A_u$ 
 \caption{\texttt{computeValuation}$(\mathcal{T},v,\mathcal{S})$ (By the Basic Algorithm)}
 \label{algo:computeValuationByBA}
\end{algorithm}

The original Basic Algorithm \cite{basicalgorithm1,basicalgorithm2}  uses heuristics for deciding a variable elimination order, instead of a project-join tree.   Algorithm \ref{algo:computeValuationByBA} can be viewed as a combination of Basic Algorithm and the project-join-tree-based approach. Note that Algorithm \ref{algo:computeValuationByBA} does not compute the sum of ADDs from the sub-valuations, though it does compute the sum of derivatives. Instead, it stores the valuations by an additive decomposition, which is a set of ADDs. Therefore, Algorithm \ref{algo:computeValuationByBA} provides the potential of parallelizing the symbolic approach.

We compare in practice, the efficiency of Algorithm \ref{algo:computeValuationByADDMax} and \ref{algo:computeValuationByBA} used in Algorithm \ref{algorithm:algorithm:ADD-Based-MaxSAT-Solving} on a subset of benchmarks used in evaluation, which contains 100 relatively small instances. We track in Table \ref{table:basicAlgorithm} the running time and the maximum number of nodes in a valuation during the computation. \footnote{In Algorithm \ref{algo:computeValuationByBA}, the number of ADD nodes in a valuation is the sum of nodes in the additive decomposition.} According to the results, Algorithm \ref{algo:computeValuationByADDMax} introduced in the main paper outperforms Algorithm \ref{algo:computeValuationByBA} based on Basic Algorithm. On average, Algorithm \ref{algo:computeValuationByADDMax} uses about $1/3$ of the running time and $1/18$ the number of nodes compared with Algorithm \ref{algo:computeValuationByBA}. Therefore, the additive decomposition used in Algorithm \ref{algo:computeValuationByBA} is generally less efficient.

	\begin{table}
		\centering
		\begin{footnotesize}
		\begin{tabular}{c c c c c}
			\toprule 
		Algorithm & & 	Alg. \ref{algo:computeValuationByADDMax} & & 	Alg. \ref{algo:computeValuationByBA} (Basic Algorithm) \\
			\cmidrule{1-1} \cmidrule{3-3} \cmidrule{5-5}
Avg. Running Time (s)  & & 1.94 && 7.53\vspace{0.1cm}   \\ 
Avg. Maximum \# Nodes     & & 	52450 && 497642 \vspace{0.1cm} \\ 
Avg. Speed-up ($\times$) compared with Alg. \ref{algo:computeValuationByBA} && 3.4 && 1\\
Avg. Reduction ($\times$) of Maximum \# Nodes
compared with Alg. \ref{algo:computeValuationByBA}  && 18.4 && 1\\
			\bottomrule
		\end{tabular} 
			\caption{\textbf{Comparison between Algorithm \ref{algo:computeValuationByADDMax} and Algorithm \ref{algo:computeValuationByBA}}. On average, Algorithm \ref{algo:computeValuationByADDMax} uses about $1/3$ of the running time and $1/18$ the number of nodes compared with Algorithm \ref{algo:computeValuationByBA}. Therefore, the additive decomposition used in Algorithm \ref{algo:computeValuationByBA} is generally less efficient.}
		\label{table:basicAlgorithm}
			\end{footnotesize}
	\end{table}
	
\section{Different Implementations of $\max$ on ADDs}
Although the performance of Algorithm \ref{algo:computeValuationByBA} is generally worse than Algorithm \ref{algo:computeValuationByADDMax}, the idea of maximization by substitution (composition) provides an alternative way of implementing $max$, i.e.,

\begin{equation}
\label{equa:ba}
\tag{B.1}
\max_xF \equiv F_{|x\gets G_x}\equiv F_{|x\gets \argmax_xF}.
\end{equation}

Instead, the $\max$ in Algorithm \ref{algo:computeValuationByADDMax} is done by 

\begin{equation}
\max_xF \equiv \max\{F_{|x\gets 1}, F_{|x\gets 0}\},
\label{equa:max}
\tag{B.2}
\end{equation}

where computing $G_x$ is not needed. 

Generally, $\max$ by Equation (\ref{equa:max}) is faster than Equation (\ref{equa:ba}), if a maximizer is not needed (computing the maximizer require the computation of $G_x$ for each variable $x$). However, when generating a maximize is necessary, then $G_x$ can be used in both variable elimination in Equation (\ref{equa:ba}) and maximizer generation, which makes Equation (\ref{equa:ba}) more efficient than Equation (\ref{equa:max}) for implementing $\max$. Therefore, in the implementation of DPMS, we use Equation (\ref{equa:max}) to implement $\max$ when a maximizer is not needed for better efficiency, and use Equation (\ref{equa:ba}) otherwise.

\section{Technical Proofs}
\subsection{Proof of Proposition \ref{prop:earlyMaximization}}
Since $\nabla_xG \equiv 0 $, we have $G(b\cup \{x\})=G(b)$ for all $b\in \mathbb{B}^n$.

Therefore, for all $b\in \mathbb{B}^X$, we have
\begin{align*}
&\mathop{\max}_x(F+G)(b)\\
=&max\{F(b\cup \{x\})+G(b\cup \{x\}),F(b)+G(b)\}  \\
=&max\{F(b\cup \{x\}),F(b)\}+G(b)\\ 
=&\mathop{\max}_xF(b)+G(b)\\
\end{align*}
Thus we have $\mathop{\max}\limits_x(F+G)\equiv \mathop{\max}\limits_xF+G$.\hfill\qedsymbol

\subsection{Proof of Theorem \ref{theo:correctness}}
\begin{pfs*}
This proof is an adaption of the proof of Theorem 2 in \cite{dpmc}. The idea is to figure out the non-recursive interpretation of the valuation $\mathcal{F}_v^{\mathcal{T}}$ (Lemma \ref{lemma:invariant}) from the recursive definition of valuation. Structural induction on the project-join tree is frequently used in the proof.
\end{pfs*}

First, we define some useful notations:

Given a project-join tree $\mathcal{T}=(T,r,\gamma,\pi)$ and a node $v$, denote by $T(v)$ the subtree rooted at $v$. We define the set $\Phi(v)$ of constraint that correspond to the leaves of $T(v)$:
\begin{align*}
    \Phi(v)\equiv
    \begin{cases}
    \{\gamma(v)\} &\text{if $v\in L(T)$}\\
        \bigcup_{u\in \texttt{children}(v)}\Phi(u)& \text{otherwise}
    \end{cases}
\end{align*}

We also define the set $P(v)$ of all variables to project in the subtree $T(v)$:

\begin{align*}
    P(v)\equiv
    \begin{cases}
    \emptyset &\text{if $v\in L(T)$}\\
        \bigcup_{u\in \texttt{children}(v)}P(u)\cup \pi(v)& \text{otherwise}
    \end{cases}
\end{align*}

Note that for the root $r$, we have $\Phi(r)=C_f$ and $P(r)=X$. The following two lemmas will be used later.

\begin{lemma}
For a tree $T$ and an internal node $v$, let $u_1,u_2$ be two distinct children of $v$. Then we have
$$
P(u_1)\cap P(u_2)=\emptyset
$$
\label{lemma:Ppartition}
\end{lemma}
\begin{proof*}
First node that the subtree with root $u_1$ and the subtree with root $u_2$ do not share nodes. In other words, we have 

$$
V(T(u_1))\cap V(T(u_2))=\emptyset
$$
This is true because otherwise we will have a loop in a tree, which is impossible.

Assume $P(u_1)\cap P(u_2)\neq\emptyset$ and there exists $x\in P(u_1)\cap P(u_2)$. Then by the definition of $P$, there exist two distinct internal nodes $o_1\in V(T(u_1))$ and $o_2\in V(T(u_2))$, such that $x\in \pi(o_1)$ and $x\in \pi(o_2)$. However, this contradicts with the property of the project-join tree:  $\{\pi(v):v\in V(T)\setminus L(T)\}$ is a partition of the variable set $X$. \hfill\qedsymbol
\end{proof*}

\begin{lemma}
For a tree $T$ and an internal node $v$, let $u_1,u_2$ be two distinct children of $v$. Then we have
$$
\Phi(u_1)\cap \Phi(u_2)=\emptyset.
$$
\label{lemma:Phipartition}
\end{lemma}

\begin{proof*}
Suppose there exists $c\in \Phi(u_1)\cap \Phi(u_2)$, then by the definition of $\Phi$, the leaf node corresponding to $c$, i.e.,  $\gamma^{-1}(c)$ must be a descendant of both $u_1$ and $u_2$, which conflicts with the fact that 
$
V(T(u_1))\cap V(T(u_2))=\emptyset,
$ proved in the proof of Lemma \ref{lemma:Ppartition}. \hfill\qedsymbol
\end{proof*}

Then, in order to prove Theorem \ref{theo:correctness}, we first propose and prove the following invariant Lemma.
\begin{lemma}
After all valuations are computed, we have for each node $v\in V(T)$, 
$$
\mathcal{F}_v^\mathcal{T} \equiv \mathop{\max}\limits_{P(v)}\sum_{c\in \Phi(v)}c
$$
\label{lemma:invariant}
\end{lemma}
\begin{proof*}
We prove Lemma \ref{lemma:invariant} by structural induction.

For the basis case, $v\in L(T)$, we have $\Phi(v)=\gamma(v)$ and $P(v)=\emptyset$.
By Definition \ref{defn:valuation}, 

$$
\mathcal{F}_v^\mathcal{T}=\gamma(v)=\mathop{\max}\limits_{P(v)}\sum_{c\in \Phi(v)}c
$$
trivially holds.

For the inductive step, i.e., $v\in V(T)\setminus L(T)$.

By induction hypothesis, for all $u\in\texttt{children}(v)$, we have 
$$
\mathcal{F}_u^\mathcal{T} \equiv \mathop{\max}\limits_{P(u)}\sum_{c\in \Phi(u)}c
$$

Then 
\begin{align*}
    &\mathcal{F}_v^\mathcal{T}\\
    \equiv&  \mathop{\max}\limits_{\pi(v)}\sum\limits_{u\in \texttt{children}(v)}\mathcal{F}_u^\mathcal{T} \text{\,\,\,\,\,\,\,(Definition \ref{defn:valuation})}\\
       \equiv&  \mathop{\max}\limits_{\pi(v)}\sum\limits_{u\in \texttt{children}(v)}\mathop{\max}\limits_{P(u)}\sum_{c\in \Phi(u)}c \text{\,\,\,\,\,\,\,(inducton hypothesis)}\\
        \equiv&  \mathop{\max}\limits_{P(v)}\sum\limits_{u\in \texttt{children}(v)}\sum_{c\in \Phi(u)}c \text{\,\,\,\,\,\,\,(by Lemma \ref{lemma:Ppartition} and undoing early maximization multiple times)}\\
           \equiv&  \mathop{\max}\limits_{P(v)}\sum_{c\in \Phi(v)}c \text{\,\,\,\,\,\,\,(Lemma \ref{lemma:Phipartition} )}\\
\end{align*}
\hfill\qedsymbol
\end{proof*}
Now we are ready to prove Theorem \ref{theo:correctness}. For the root $r$ of the project-join tree, we have
\begin{align*}
    &\mathcal{F}_r^\mathcal{T}
    =\mathop{\max}\limits_{P(r)}\sum_{c\in \Phi(r)}c
    =\mathop{\max}\limits_{X}\sum_{c\in C_f}c=\mathop{\max}\limits_{X}F_f
\end{align*}
\hfill\qedsymbol

\subsection{Proof of Proposition \ref{prop:maximizer}}
    \begin{align*}
        &F(\{\argmax_xF(b^\star)\}\cup b^\star)\\
        =&\max\{F(b^\star\cup \{x\}),F(b^\star)\}\\
        =&\max_xF(b^\star)\\
        =&\max_{X\setminus\{x\}}\max_x F\\
        =&\max_XF
    \end{align*}

\subsection{Proof of Proposition \ref{prop:argmaxAndDerivative}}
This proposition can be easily proved by observing that $F({b}\cup \{x\})\ge F({b})$ is equivalent with $\nabla_xF(b)\ge 0$ by Definition \ref{defn:optimization} and \ref{defn:derivative}.\hfill\qedsymbol

\subsection{Proof of Theorem \ref{theo:maximizerAlg}} 
\begin{pfs*}
The proof keeps track of the ``active" ADDs during the computation. To do this, we slightly change Algorithm \ref{algo:computeValuationByADDMax} to have a set $Q$ to contain ``active" ADDs. Then we show that, suppose at a moment variables in $S$ have been eliminated, the sum of currently active ADDs equals the pseudo-Boolean function $max_SF_f$. As a result, the $G_x$ computed in Algorithm \ref{algo:computeValuationByADDMax} is the $\argmax_x$ of $max_SF_f$, which is what we need to extend a maximizer by appending the value of $x$. When the last variable $x_l$ is eliminated, $G_{x_l}$   will be a length-$1$ maximizer containing the value of $x_l$. In Algorithm \ref{algo:maximizer}, a maximizer is iteratively built from length-$1$ to length-$|X|$ by the reverse order of variable, which is enforced by  the stack $\mathcal{S}$. For example, $x_l$ will be the first variable popped in Algorithm \ref{algo:maximizer}.
\end{pfs*}

For a formula $f$, recall the objective function is defined as
$$
F_f({b})= \sum_{c\in C_f}{c({b})}, 
$$
for all $b\in \mathbb{B}^{X}$.

\begin{algorithm}[h]
    \SetAlgoLined
    \SetKwInOut{Input}{Input}
    \SetKwInOut{Output}{Output}
    \Input{A project-join tree $\mathcal{T}=(T,r,\gamma,\pi)$, a node $v\in V(T)$, a stack $\mathcal{S}$ for constructing the maximizer.}
    \Output{An ADD representing of the valuation of $v$, i.e., $\mathcal{F}^\mathcal{T}_v$}
    \vspace{0.1cm}
    \hrule
    \vspace{0.1cm}
   \lIf{$v\in L(T)$ } 
    {\Return $\texttt{compileADD}(\gamma(v))$\label{line:algo2line1}} 
    $A\gets \texttt{ZeroADD}$\\
    \For{$u\in \texttt{children}(v)$}{
        $A'=\texttt{computeValuation}(\mathcal{T},u,\mathcal{S})$\\
        $Q.remove(A')$ \label{line:algo5line5}\\
        $A\gets A +A'$\\
    }
    $Q.insert(A)$\label{line:algo5line7}\\
      \For{$x\in \pi(v)$}{
            \text{ \texttt{/* For maximizer generation */} } \label{line:algo5line9}\\
        $G_x\gets \texttt{sgn}(\nabla_xA)$ \label{line:algo5line10}\\
    $\mathcal{S}$.\texttt{push}$((x,G_x))$\label{line:algo5line11}\\
      \text{ \texttt{/* Eliminate $x$ */} }\\
      $Q.remove(A)$\label{line:algo5line13}\\
    $A\gets \mathop{\max}_{x}A$ \label{line:algo5line14}\\ 
    $Q.insert(A)$\label{line:algo5line15}\\
         }
  \Return $A$ 
 \caption{\texttt{computeValuation}  $(\mathcal{T},v,\mathcal{S})$ (modified, for proof of Lemma \ref{lemma:QisObjective} )}
 \label{algo:twistedcomputeValuation} 
\end{algorithm}

In order to complete this proof, we slightly modified Algorithm \ref{algo:computeValuationByADDMax} to Algorithm \ref{algo:twistedcomputeValuation}. Compared with Algorithm \ref{algo:computeValuationByADDMax}, Algorithm \ref{algo:twistedcomputeValuation} uses a set $Q$ of ADDs to keep track of the ``active" valuations. Initially, $Q$ is the set of the ADDs of all  constraints. Despite of the set $Q$, Algorithm \ref{algo:twistedcomputeValuation} computes the same valuation with Algorithm \ref{algo:computeValuationByADDMax}. During the execution of Algorithm \ref{algo:twistedcomputeValuation}, we have the following lemma, saying that the sum of ADDs in $Q$ is the ADD representation of the pseudo-Boolean function $\max_SF_f$, where $S$ is the set of variables that already have been eliminated.

\begin{lemma}
In Algorithm \ref{algo:twistedcomputeValuation}, before Line \ref{line:algo5line11},  a variable-ADD pair $(x,G_x)$ is pushed into the stack $\mathcal{S}$, suppose the variable-ADD pairs for variables in $S$ ($S\subseteq X$ and $x\not \in S$) have been pushed into the stack $\mathcal{S}$, then we have $$\sum_{D\in Q}D=\max_SF_f.$$ 
\label{lemma:QisObjective}
\end{lemma}

\begin{proof*} We prove by induction with the order of variable elimination.

Basis case: Before the first variable is pushed, $\sum_{A\in Q}=\sum_{c\in C_f}=F_f$ by initialization of $Q$.

For the inductive step, suppose in the iteration where $x$ will be eliminated, while the previous variable eliminated is $x'$. By induction hypothesis, before $x'$ is eliminated, we have

$$
\sum_{D\in Q}D=max_{S\setminus\{x\}}F_f
$$

Between the moment before executing Line \ref{line:algo5line11} in the iteration of $x'$ and  Line \ref{line:algo5line11} in the iteration of $x$, the set $Q$ may be changed by:

\begin{enumerate}
    \item Line \ref{line:algo5line13} and \ref{line:algo5line15} in the iteration of $x'$, where $A$ is replaced by $max_{x'}A$ in $Q$. Since $A$ is the only ADD in $Q$ that contains $x$, after the replacement, we have 
    $$
    \sum_{D\in Q}D=max_{S}F_f.
    $$
    \item Line \ref{line:algo5line5} and \ref{line:algo5line7} in Algorithm \ref{algo:twistedcomputeValuation}, where several ADDs in $Q$ is replaced by their sum (note that when an ADD is removed from $Q$, it must have been inserted into $Q$ Line \ref{line:algo5line15} in the recursive call to \texttt{computeValuation}). However, this operation does not change the pseudo-Boolean function $\sum_{D\in Q}D$.
\end{enumerate}
Therefore before $x$ is eliminated, we have $\sum_{D\in Q}D=max_SF_f.$\hfill\qedsymbol

\end{proof*}

We first prove the following loop invariants regarding Algorithm \ref{algo:computeValuationByADDMax} and \ref{algo:maximizer}.

\begin{lemma}
\label{lemma:algorithm2}
In Algorithm \ref{algo:computeValuationByADDMax}, after Line \ref{line:algo2line5} is executed, i.e., $G_x$ is computed, suppose the variable-ADD pairs for variables in $S$ ($S\subseteq X$ and $x\not \in S$) have been pushed into the stack $\mathcal{S}$, then we have $$G_x=\argmax_x(\max_SF_f).$$ 
\label{lemma:Gx}
\end{lemma}

\begin{proof*} 
Note that $G_x$ in Algorithm \ref{algo:computeValuationByADDMax} and \ref{algo:twistedcomputeValuation} is equivalent. Therefore, we prove the result for Algorithm \ref{algo:twistedcomputeValuation} and it also holds for Algorithm \ref{algo:computeValuationByADDMax}. 

According to Lemma \ref{lemma:QisObjective}, we know that before Line \ref{line:algo5line10} of Algorithm \ref{algo:twistedcomputeValuation}, the following holds:

$$
\sum_{D\in Q}D=max_SF_f
$$
Since $\mathcal{T}$ is a project-join tree and $x$ will be eliminated in this iteration, $A$ is the only ADD in $Q$ that contains $x$. Therefore 

$$G_x=\texttt{sgn}(\nabla_xA)=\texttt{sgn}(\nabla_x \max_SF_f)$$

By Proposition \ref{prop:argmaxAndDerivative}, $G_x$ is the ADD representation of $\argmax_x (max_SF_f)$. \hfill\qedsymbol

\end{proof*}

\begin{lemma}
When a variable-ADD pair $(x,G_x)$ is popped in Algorithm \ref{algo:maximizer},  suppose the variable-ADD pairs for variables in $S$ ($S\subseteq X$ and $x\not \in S$) are still in the stack $\mathcal{S}$, then after the execution of Line \ref{line:algo3line4} of Algorithm \ref{algo:maximizer}, we have $b^\star$ is a maximizer of $\max_{S}F_f$.
\label{lemma:algorithm3}
\end{lemma}
\begin{proof*}
We prove by induction on the loop of Algorithm \ref{algo:maximizer}. 

Basis case: Before entering the loop, $b^\star=\emptyset$ and $S=X$. The $b^\star$, i.e., the  empty set is a trivial maximizer of the scalar $max_XF_f$.

Inductive step: Consider the iteration where $(x,G_x)$ is popped. By inductive hypothesis, the $b^\star$ before Line \ref{line:algo3line4} is a maximizer of $max_{S\cup\{x\}}F_f=max_xmax_SF_f$.  After $(x,G_x)$ is popped, by Lemma  \ref{lemma:algorithm2}, we have $G_x=\argmax_x(\max_SF_f)$. By Proposition \ref{prop:maximizer}, we have $b^\star \cup G_x(b^\star)=b^\star \cup \{\argmax_x(\max_SF_f)(b^\star)\}$ is a maximizer of $max_SF_f$.\hfill\qedsymbol
\end{proof*}

Now we are able to prove Theorem \ref{prop:maximizer}. Given Lemma \ref{lemma:algorithm3}, Theorem \ref{prop:maximizer} directly follows by consider the last variable popped from the stack: $S=\emptyset$ and $b^\star$ is a maximizer of $F_f$. \hfill\qedsymbol

\subsection{Proof of Theorem \ref{theo:treewidth}}
\begin{pfs*}
We first formally define the width of a (graded) project-join tree and the size of a formula. Then we analyze each operation of Algorithm  \ref{algorithm:algorithm:ADD-Based-MaxSAT-Solving} and prove again by structural induction that the running time of all three types of operations is bounded by $O(k\cdot 2^k\cdot |f|)$.
\end{pfs*}
To prove this theorem, we first define the $width$ of a (graded) project-join tree formally.
For each constraint $c \in C(f)$, define \texttt{Vars}$(c)$ to be the set of variables appearing in $c$.

For each node $v\in V(T)$, $\texttt{Vars}(v)$ is defined as follows: 

\begin{align*}
    \texttt{Vars}(v)\equiv
    \begin{cases}
    \texttt{Vars}(\gamma(v)) &\text{if $v\in L(T)$}\\
        \big(\bigcup_{u\in \texttt{children}(v)}\texttt{Vars}(u)\big)\setminus \pi(v)& \text{otherwise}
    \end{cases}
\end{align*}

The size of a node $v$, $size(v)$ is defined to be $|\texttt{Vars}(v)|$ if $v$ is a leaf, and $|\texttt{Vars}(v)\cup \pi(v)|$ if $v$ is an internal node.

The \emph{width} of a (graded) project-join tree $\mathcal{T}=(T,r,\gamma,\pi)$, denoted by $width(\mathcal{T})$, is defined as

$$
width(\mathcal{T}) \equiv \max_{v\in V(T)} size(v).
$$

For a node $v$ in the project-join tree, we define the sub-formula associated with $v$, denoted by $f_v$ as follows:

\begin{align*}
    f(v)\equiv
    \begin{cases}
    \gamma(v) &\text{if $v\in L(T)$}\\
        \bigwedge_{u\in \texttt{children}(v)}f_u& \text{otherwise}
    \end{cases}
\end{align*}

For a conjunctive formula $f$, we define the size of $f$, denoted by $|f|$, by the sum of the number of variables in all constraints.

We prove the following statement by structural induction:
\begin{lemma}
For a  (graded) project-join tree $\mathcal{T}=(T,r,\gamma,\pi)$ and a node $v\in V(T)$, if $width(\mathcal{T})\le k$, then \texttt{computeValuation}$(\mathcal{T},v,\mathcal{S} )$ terminates in $O(k2^k|f_u|)$. 
\end{lemma}
\begin{proof*} We prove this lemma by structural induction.

For the basis case, $v\in L(T)$, we have $f_v=$\texttt{compileADD}$(\gamma(v))$ and the running time of \texttt{computeValuation} is that of $\texttt{compileADD}(\gamma(v))$. Since $width(\mathcal{T})\le k$, we know that $\gamma(v)$ has at most $k$ variables. Assume that given an assignment, the value of a constraint can be determined in $O(|\gamma(v)|)$, which is true for most types of constraints. Then $\texttt{compileADD}(\gamma(v))$ can be done in $O(k\cdot 2^k \cdot |f_v|)$ by simply building the binary decision tree $(O(2^k\cdot |f_v|))$ and reduce it ($O(k\cdot 2^k)$). 

For the inductive step, i.e., $v\in V(T)\setminus L(T)$.

By induction hypothesis, for all $u\in\texttt{children}(v)$, we have \texttt{computeValuation}$(\mathcal{T},u,\mathcal{S} )$ terminates in $O(k2^k|f_u|)$.

Next we analyze the computation done in Algorithm \ref{algo:computeValuationByADDMax}.   The overall observation is that  the number of variables in the ADD $A$ during the execution of the algorithm is always bounded by $k$, given $width(\mathcal{T})\le k$. Therefore the size of an intermediate (reduced, ordered) ADD  is always bounded by $O(2^k)$.
\begin{enumerate}
    \item  Line \ref{line:algo2line2} calls \texttt{computeValuation} on all children of $v$. By the induction hypothesis, the total time is bounded by
    $$
    \sum_{u\in \texttt{children}(v)}O(k2^k|f_u|)=O(k2^k|f_v|)
    $$
Note that $|f_v|= \sum_{u\in \texttt{children}(v)}|f_u|$ by Lemma \ref{lemma:Phipartition}.

\item Line \ref{line:algo2line2} also sums up all intermediate ADDs provided by its children. Since the $width$ of $\mathcal{T}$ is bounded by $k$, the number of variables of two ADDs for summing are at most $k$. Thus sum can be done in $O(k\cdot 2^k)$ by simply enumerating all possible $2^k$ assignments, building a Boolean decision tree and reduce it. Therefore, the total time is bounded by $O(|\texttt{children}(v)|\cdot k\cdot 2^k)$. Note that $|\texttt{children}(v)|= \sum_{u\in \texttt{children}(v)}1\le \sum_{u\in \texttt{children}(v)}|f_u|=|f_v|$. Thus the total time is also bounded by $O(k\cdot 2^k|f_v|)$.

\item Line \ref{line:algo2line5} and \ref{line:algo2line6} push $\texttt{sgn}(\nabla_xA)$ to the stack for generating the maximizer. The complexity of computing $\nabla_x A$ and \texttt{sgn}$(\nabla_xA)$ are both $O(k\cdot 2^k)$. Since variables in $\pi(v)$ are eliminated,  the total time is bounded by $O(|\pi(v)|\cdot k\cdot 2^k)$. We have
$$|\pi(v)|\le \big|\bigcup_{u\in \texttt{children}(v)}\texttt{Vars}(f_u)\big|\le \sum_{u\in \texttt{children}(v)}|\texttt{Vars}(f_u)|\le \sum_{u\in \texttt{children}(v)}|f_u|=|f_v|.$$
 Thus the total time is also bounded by $O(k\cdot 2^k|f_v|)$.
\end{enumerate}

Hence, the total time of $1$-$3$ above is still bounded by $O(k\cdot 2^k \cdot |f_v|)$. \hfill\qedsymbol
\end{proof*}

We are ready to complete the proof of Theorem \ref{theo:treewidth}. By Lemma 2, \texttt{computeValuation}$(\mathcal{T},r,\mathcal{S} )$ terminates in $$O(k\cdot 2^k|f_r|)=k\cdot 2^k|f|$$.

Finally, the running time of  Algorithm \ref{algo:maximizer} for building the maximizer is bounded by $ O(n\cdot k)=O(|f|\cdot k)$.

Therefore, the total running time of Algorithm \ref{algorithm:algorithm:ADD-Based-MaxSAT-Solving} is bounded by $O(k\cdot 2^k\cdot |f|)$. \hfill\qedsymbol

\end{appendix}
\end{document}